\PassOptionsToPackage{table,xcdraw,usenames,dvipsnames}{xcolor}
\documentclass{article} % For LaTeX2e
\usepackage{iclr2025_conference,times}

\usepackage{amsmath,amsfonts,bm}

\def\eqref#1{equation~\ref{#1}}
\def\1{\bm{1}}

\DeclareMathAlphabet{\mathsfit}{\encodingdefault}{\sfdefault}{m}{sl}
\SetMathAlphabet{\mathsfit}{bold}{\encodingdefault}{\sfdefault}{bx}{n}

\usepackage{hyperref}
\usepackage{url}
\usepackage{amsthm}
\usepackage{hyperref}

\usepackage[most]{tcolorbox}
\usepackage{wrapfig}
\usepackage{graphicx,float}
\usepackage{subfigure}
\usepackage{threeparttable}
\usepackage[ruled,vlined,noend]{algorithm2e}
\usepackage{colortbl}
\usepackage{color}
\usepackage{multirow}
\usepackage{tabularx}
\usepackage{float}
\usepackage{graphicx}
\usepackage{booktabs}
\usepackage{arydshln}
\usepackage{enumitem}
\usepackage{wrapfig}
\usepackage{caption}
\usepackage{graphicx}
\usepackage{makecell}
\usepackage{float} 
\usepackage{mathtools}
\usepackage{subfigure}
\usepackage{bbding}
\usepackage{makecell}
\usepackage{tabularx}
\usepackage{amssymb,mathrsfs,amsmath}
\usepackage{pifont}
\usepackage{bbm}
\usepackage{siunitx}

\usepackage{listings}

\newcommand{\ourmethod}{{\fontfamily{lmtt}\selectfont \textbf{AgenTracer}}\xspace}
\newcommand{\ourmodel}{{\fontfamily{lmtt}\selectfont \textbf{AgenTracer}-8B}\xspace}
\newcommand{\ourdata}{{\fontfamily{lmtt}\selectfont \textbf{TracerTraj}}\xspace}

\definecolor{bittersweet}{rgb}{1.0, 0.44, 0.37}
\definecolor{mygreen}{rgb}{0.29, 0.7, 0.48}
\usepackage{pifont}% http://ctan.org/pkg/pifont for xmark and cmark
\definecolor{demphcolor}{RGB}{144,144,144}

\definecolor{mygray}{gray}{0.4}
\definecolor{autopurple}{HTML}{7030A0}
\definecolor{dyna_yellow}{HTML}{BF9000}
\definecolor{adaptive_blue}{HTML}{0070C0}
\definecolor{darksalmon}{rgb}{0.91, 0.59, 0.48}
\definecolor{emerald}{rgb}{0.31, 0.78, 0.47}
\definecolor{green(pigment)}{rgb}{0.0, 0.65, 0.31}
\definecolor{amaranth}{rgb}{0.9, 0.17, 0.31}
\definecolor{iris}{rgb}{0.35, 0.31, 0.81}
\definecolor{uu}{rgb}{0.95, 0.51, 0.51}
\definecolor{spirodiscoball}{rgb}{0.06, 0.75, 0.99}
\usepackage{svg}

\usepackage{fontawesome5}
\usepackage{twemojis}

\usepackage{cleveref}
\SetKwInOut{Input}{Input}\SetKwInOut{Output}{Output}

\SetCommentSty{mycommfont}
\SetKwComment{Comment}{$\triangleright$ }{}

\definecolor{mygrey}{gray}{0.4}

\hypersetup{
    colorlinks=true,
    linkcolor=red,
    citecolor=cyan,
    filecolor=magenta,      
    urlcolor=magenta,
    }

\title{\includegraphics[width=0.05\textwidth]{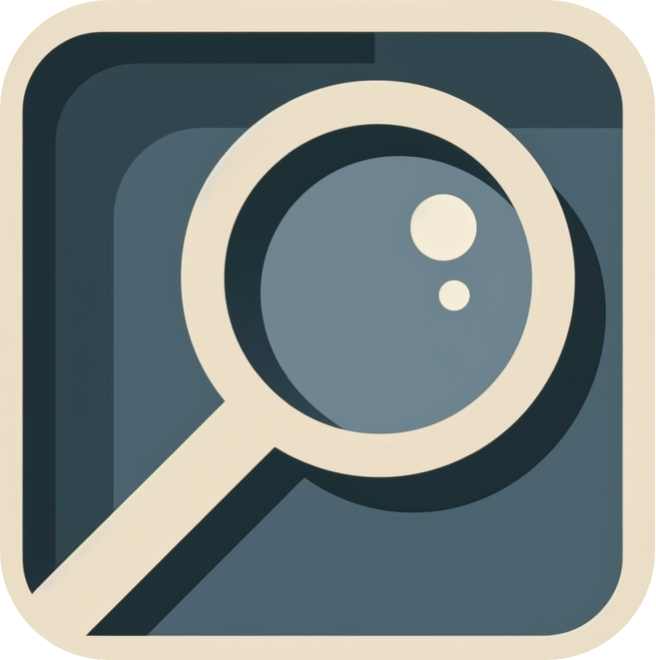}AgenTracer: Who Is Inducing Failure in the LLM Agentic Systems?}

\author{Guibin Zhang$^{\twemoji{lion}\dag}$, 
Junhao Wang$^{\twemoji{bird}\dag}$, 
Junjie Chen$^{\twemoji{lion}\dag}$, \textbf{Wangchunshu Zhou}$^{\twemoji{evergreen tree}}$,\\
\textbf{Kun Wang}$^{\twemoji{deer}\twemoji{envelope}}$, 
\textbf{Shuicheng Yan}$^{\twemoji{lion}\twemoji{envelope}}$ \\
	$^{\twemoji{lion}}$NUS\quad
    $^{\twemoji{bird}}$CUHK \quad
    $^{\twemoji{evergreen tree}}$OPPO\quad
    $^{\twemoji{deer}}$NTU \\
{\faEnvelope} {Main Contact}: \texttt{guibinz@outlook.com}
}

\iclrfinalcopy 
\begin{document}

\maketitle

\begin{abstract}
Large Language Model (LLM)-based agentic systems, often comprising multiple models, complex tool invocations, and orchestration protocols, substantially outperform monolithic agents. Yet this very sophistication amplifies their fragility, making them more prone to system failure. Pinpointing the specific agent or step responsible for an error within long execution traces defines the task of \textbf{agentic system failure attribution}. Current state-of-the-art reasoning LLMs, however, remain strikingly inadequate for this challenge, with accuracy generally below $10\%$.  
To address this gap, we propose \ourmethod, the first automated framework for annotating failed multi-agent trajectories via counterfactual replay and programmed fault injection, producing the curated dataset \ourdata. Leveraging this resource, we develop \ourmodel, a lightweight failure tracer trained with multi-granular reinforcement learning, capable of efficiently diagnosing errors in verbose multi-agent interactions. On {Who\&When} benchmark, \ourmodel outperforms giant proprietary LLMs like Gemini-2.5-Pro and Claude-4-Sonnet by up $18.18\%$, setting a new standard in LLM agentic failure attribution. More importantly, \ourmodel delivers actionable feedback to off-the-shelf multi-agent systems like MetaGPT and MaAS with $4.8\sim14.2\%$ performance gains, empowering self-correcting and self-evolving agentic AI. Our project page is at \url{https://bingreeky.github.io/atracer/}.

\end{abstract}

\begin{figure*}[!h]
\centering
\includegraphics[width=\linewidth]{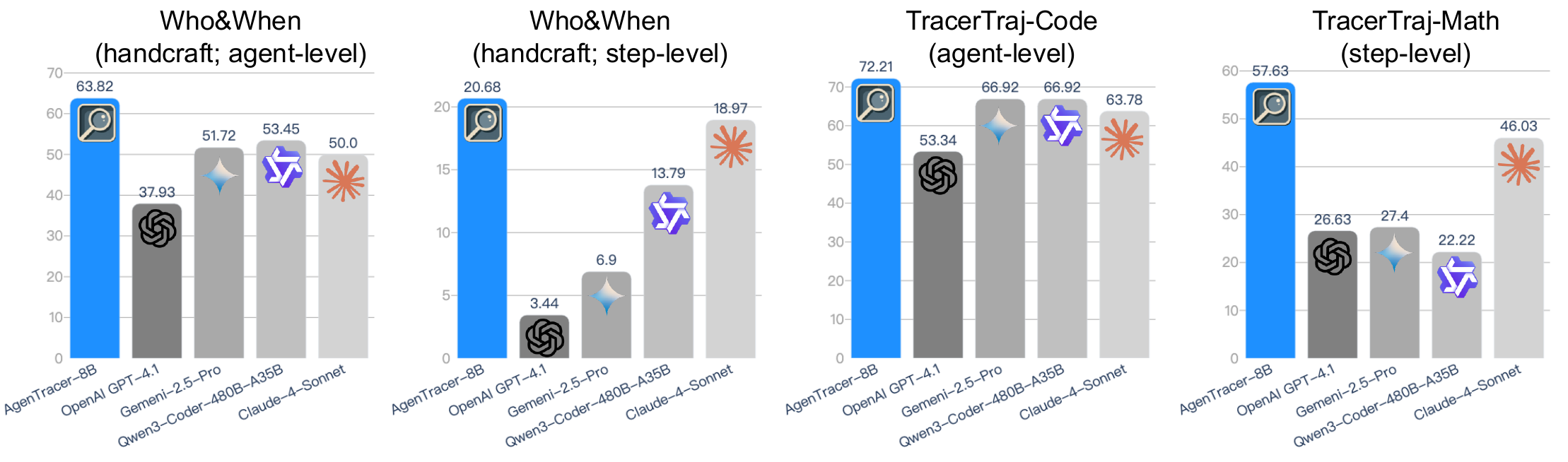}
\vspace{-1.8em}
\caption{\small Benchmark performance comparison between \ourmodel and leading industry providers.}
\label{fig:main-perf}
\vspace{-1.3em}
\end{figure*}

\section{Introduction}
{Large Language Model (LLM)-powered agents} have exhibited exceptional proficiency across a wide array of cognitive faculties, encompassing perception~\citep{driess2023palm-e,wang2024omnidrive,zheng2023steve,wei2024editable}, planning~\citep{zhu2024knowagent,erdogan2025plan-and-act,huang2024understanding}, reasoning~\citep{putta2024agentq,masterman2024landscape}, and action~\citep{li2024embodied,yang2024embodied}. As endeavors to address increasingly intricate challenges, such as iterative tool calling, complex document analysis, and multi-step web navigation, continue to surge, the inherent limitations of relying on a single monolithic model have become increasingly conspicuous.

Consequently, numerous multi-agent frameworks, motivated by collective intelligence~\citep{wang2025improvingmodelalignmentcollective,zhang2024cut,zhang2024g-designer} and the society of mind theory~\citep{Book1988_SoM,NeurIPS2023_Agent-SoM}, have emerged, ensembling multiple LLM agents to achieve refined subtask orchestration~\citep{zhang2025agentorchestrahierarchicalmultiagentframework,hu2025owloptimizedworkforcelearning}, ultra-long context handling~\citep{zhang2024chainagentslargelanguage}, and broader environmental perception~\citep{jiang2024multiagentvqaexploringmultiagent}. These systems have demonstrated superior performance over single-agent counterparts across a range of complex real-world domains, including data science engineering~\citep{hong2024data-interpreter}, scientific discovery~\citep{ghareeb2025robinmultiagentautomatingscientific,ghafarollahi2024sciagentsautomatingscientificdiscovery}, and software engineering~\citep{wei2025swerladvancingllmreasoning}. However, the integration of multiple autonomous agents alongside diverse auxiliary modules (\textit{e.g.}, external databases, tools, and memory modules) inevitably exacerbates \textbf{system fragility}. As evidence, a recent empirical study by UC Berkeley~\citep{cemri2025mas-fail} reveals alarmingly high failure rates (reaching up to $86.7\%$) in popular multi-agent frameworks such as OpenHands~\citep{wang2025openhandsopenplatformai} and MetaGPT~\citep{meta-gpt}, with failure modes ranging from improper task decomposition to role disobedience. Such a high incidence of failure casts a long shadow over the real-world reliability of multi-agent systems.

A natural response to this challenge is \textbf{failure attribution}, \textit{i.e.}, the precise identification of faulty components within the system upon task failure. Accurate failure attribution plays a critical role across multiple dimensions: \textbf{(I) system debugging}, by enabling agentic systems to perform more effective self-debugging across iterative attempts, thereby enhancing performance; \textbf{(II) data efficiency}, by leveraging failed agent trajectories to construct more informative training data; \textbf{(III) grounded self-improvement}, by facilitating grounded self-correction through precise agent credit assignment.
Despite its importance, this process is predominantly left as a manual endeavor, requiring considerable human effort to analyze verbose trajectory logs. 

While preliminary attempts have been made toward automation, their efficacy remains limited. For instance, \citet{zhang2025agent-who-and-when} employed state-of-the-art models such as OpenAI-o1 and DeepSeek-R1~\citep{deepseekai2024deepseekv3technicalreport} to perform failure attribution over trajectories from GAIA~\citep{mialon2023gaia}, yet achieved accuracy below $10\%$.
More critically, existing benchmarks for agentic failure attribution remain notably constrained. To the best of our knowledge, MAST~\citep{cemri2025mas-fail} and Who\&When~\citep{zhang2025agent-who-and-when} contain only 200 and 127 manually annotated multi-agent failure trajectories, respectively, offering limited scale for systematic evaluation. Accordingly, substantial research gaps remain along two critical axes: \ding{182} \textbf{training resources}, concerning the automated construction of large-scale annotated multi-agent trajectories; and \ding{183} \textbf{methodology}, on developing swift and accurate multi-agent failure localizer, which we also refer to as a \textit{tracer}.

\vspace{-0.5em}
\paragraph{Present Framework.} To address the aforementioned research gap, we present \textbf{(1)} \ourmethod, a fully automated pipeline for constructing well-annotated multi-agent failure trajectories,
%spanning across mathematical reasoning, code generation, web browsing, and other general-purpose agentic tasks. \ourmethod curates over 2,000 annotated failure trajectories and uses them to train
and \textbf{(2)} \ourmodel, a lightweight failure attributor for multi-agent systems. Methodologically, \ourmethod leverages \ding{168} \textbf{counterfactual replay}, systematically replacing agent actions with oracle guidance to identify the decisive error step responsible for system failure. To enhance data diversity and ensure annotation precision, it further adopts \ding{171} \textbf{programmatic fault injection}, synthetically generating failure instances by perturbing successful trajectories through targeted corruption.

Further, we fine-tune \textsc{Qwen3-8B} on the curated dataset to obtain \ourmodel via reinforcement learning (RL), enabling it to analyze long-horizon multi-agent collaboration traces. \ourmodel takes as input the trajectory log and associated environmental feedback, and outputs the identified decisive error step. A \textbf{multi-granular reward} is designed to supervise RL training, emphasizing both \textit{step-level} and \textit{agent-level} attribution accuracy.
During inference, \ourmodel enables any malfunctioning agentic system to rapidly identify the critical failure step along with explanations, thereby facilitating automated multi-agent debugging.

Our contributions are as follows:

\vspace{-0.5em}
\begin{itemize}[leftmargin=2em,itemsep=-0.1em]
\item[\ding{182}] \textbf{Automated Pipeline.} We propose {\ourmethod}, the first automated pipeline for annotating multi-agent system failures, curating over 2,000 high-fidelity failure trajectories across six datasets through counterfactual replay and programmatic fault injection.

\item[\ding{183}] \textbf{Failure Tracer.} We develop {\ourmodel}, a lightweight failure tracer dedicated for LLM agentic systems, trained via a multi-granular reinforcement learning to ensure accurate attribution at both the step and agent levels, facilitating automatic agentic system debugging.

\item[\ding{184}] \textbf{Empirical Evaluation.} Experiments show that {\ourmodel} facilitates \textit{\textbf{(I) accurate attribution}}, outperforming giant LLMs like \textsc{DeepSeek-R1} by $\sim12.21\%$ and \textsc{Gemini-2.5-Pro} by $\sim18.18\%$ on Who\&When benchmark; and \textit{\textbf{(II) self-evolution}}, enabling off-the-shelf agentic systems to improve performance by $4.8\sim14.2\%$.
\end{itemize}

\vspace{-0.5em}
\section{Related Works}
\vspace{-0.5em}
\paragraph{LLM-based Multi-Agent System.} Contemporary multi-agent systems can be broadly categorized by their level of automation into three classes:
\ding{110} \textbf{Handcrafted}, where the entire system configuration (\textit{e.g.}, LLM backbone, prompting strategies, and communication protocols) is manually specified represented by AutoGen~\citep{autogen}, AutoGPT~\citep{autogpt}, Camel~\citep{NeurIPS2023_Agent-SoM}, and ChatDev~\citep{software-dev};
\ding{110} \textbf{Partially Automated}, which automate specific system components: for example, AutoAgent~\citep{arXiv2023_AutoAgents}, LLMSelector~\citep{chen2025optimizingmodelselectioncompound}, and MasRouter~\citep{yue2025masrouter} automate agent role assignment; DsPy~\citep{khattab2023dspy} and TextGrad~\citep{yuksekgonul2024textgradautomaticdifferentiationtext} optimize prompt design; GPTSwarm~\citep{zhuge2024gptswarm} and G-Designer~\citep{zhang2024g-designer} adaptively construct inter-agent topologies;
\ding{110} \textbf{Fully Automated}, where all modules within the system are autonomously designed and evolved~\citep{hu2024adas,zhang_aflow_2024,zhang2025maas,wu2025optimas,nie2025weakforstrongtrainingweakmetaagent,gao2025flowreasonerreinforcingquerylevelmetaagents,zhang2025evoflow}.
Our method is designed to provide precise error tracing across this entire spectrum of agentic systems. Accordingly, we curate trajectories sampled from frameworks spanning all three categories.

\vspace{-0.5em}
\paragraph{Failure Attribution for Agents.} As multi-agent systems become increasingly intricate, incorporating multiple intelligent agents~\cite{multi-persona,chen2023agentverse}, tool integrations~\cite{shen2024smallllmsweaktool}, and communication protocols~\cite{marro2024scalablecommunicationprotocolnetworks}, the resulting high error rates and structural fragility have emerged as critical concerns. MAST~\citep{cemri2025mas-fail} was the first to alarmingly characterize this issue, identifying fourteen prevalent failure patterns ranging from task disobedience to reasoning-action mismatches. Building upon this, Who\&When~\citep{zhang2025agent-who-and-when} explored the feasibility of automating failure attribution, only to reveal that even top-performing reasoning models like DeepSeek-R1 fail catastrophically in this setting.
\ourmethod advances the field by introducing both a scalable data synthesis pipeline and a lightweight, high-accuracy failure attributor.

\vspace{-0.5em}
\paragraph{LLM-as-a-Judge \& Credit Assignment.} Two other research topics closely related to this work are:
\textbf{(I) LLM-as-a-Judge (LaaJ)}, which leverages LLMs/agents as evaluators based on pre-defined criteria for tasks such as data annotation~\citep{latif2025canllmaidinannotating,ChatGPT-OpenAI}, value alignment~\citep{ji2025pkusaferlhfmultilevelsafetyalignment}, reward modeling~\citep{mahan2024generativerewardmodels,lambert2024rewardbenchevaluatingrewardmodels}, and synthetic data generation~\citep{sengupta2025magvmultiagentframeworksynthetic,hu2025owloptimizedworkforcelearning}. However, when applied to multi-LLM systems, LaaJ has shown limited effectiveness, as demonstrated in~\cite{zhang2025agent-who-and-when}; \textbf{(II) Credit Assignment} is a longstanding topic in multi-agent reinforcement learning (MARL), aiming to associate individual agent actions with their long-term outcomes~\citep{pignatelli2023surveycreditassignment}. Common approaches include heuristics based on temporal contiguity~\citep{sutton1988learning}, value decomposition~\citep{arjona2019rudder}, and meta-learning~\citep{yin2023distributionalmegtagradient}. However, this problem remains largely unexplored in the context of LLM-based multi-agent systems. The most relevant prior effort, CollabUIAgents~\citep{he2025collabuiagent}, relies on LLMs to generate binary scalar ($0/1$) rewards after each agent interaction, whose reliability is inherently questionable. In contrast, \ourmethod implicitly achieves grounded credit assignment for LLM agents through principled failure attribution.

\vspace{-0.5em}
\section{Preliminary}
\vspace{-0.5em}

In this section, we provide a general definition of LLM-based multi-agent systems and their operational workflow, and then formally define the objective of the multi-agent failure attribution.

\vspace{-0.3em}
\paragraph{Notations}
\vspace{-0.5em}
Consider a LLM-based multi-agent system \( \mathcal{M} \), tasked with collaboratively resolving a user-issued query \( \mathcal{Q} \). The system consists of \( N \) agents, indexed by \( \mathcal{I} = \{1, 2, \ldots, N\} \), operating in discrete time under a turn-based protocol, \textit{i.e.}, only one agent is active at each time step. Formally:
\begin{equation}
\mathcal{M} = \bigl\langle \mathcal{I}, \mathcal{S}, \mathcal{A}, \Psi, \mu \bigl\rangle,
\end{equation}
where \( \mathcal{S} \) denotes the set of system states; \( \mathcal{A} \) is the overall action space, with each agent \( i \in \mathcal{I} \) having a local space \( \mathcal{A}_i \subseteq \mathcal{A} \); \( \mu(t) \in \mathcal{I} \) specifies the agent scheduled to act at time \( t \); \( \Psi(s_{t+1} \mid s_t, a_t, \mu(t)) \) models the state transition dynamics given the current state \( s_t \), action \( a_t \in \mathcal{A}_{\mu(t)} \), and the acting agent.
At each step \( t \), the active agent \( \mu(t) \) selects an action \( a_t \in \mathcal{A}_{\mu(t)} \) according to its policy \( \pi_{\mu(t)} \), conditioned on the current state \( s_t \), the query \( \mathcal{Q} \), and a subset of the prior interaction history \( \mathcal{H}_t \):
\begin{equation}
a_t = \pi_{\mu(t)}(s_t, \mathcal{H}_t, \mathcal{Q}),\quad \mathcal{H}_t \subseteq \{a_0, a_1, \ldots, a_{t-1}\}.
\end{equation}
The structure of \( \mathcal{H}_t \) is implementation-dependent. In LLM Debate-style frameworks~\citep{arXiv2023_MultiAgent-Debate}, it comprises the prior-round outputs from all agents; whereas in software development systems~\citep{software-dev,hu2024evomac}, a tester agent may condition only on the latest code snippet submitted by a programmer agent. The full execution trajectory of the system is denoted by:
\begin{equation}
\tau = (s_0, a_0, s_1, a_1, \ldots, s_T),
\end{equation}
where \( T \) indicates the terminal step or stopping condition. The final response to the query \( \mathcal{Q} \) is determined by the complete trajectory \( \tau \), encapsulating the collaborative behavior of all agents.

\vspace{-0.6em}
\paragraph{Objective Formulation}
A trajectory may contain multiple suboptimal actions or minor deviations, but for effective, targeted debugging, it is crucial to distinguish these from the pivotal error that renders the final outcome incorrect. Following \citep{zhang2025agent-who-and-when}, we formalize this
pivotal error as the \textbf{decisive error}, \textit{i.e.}, the earliest action in the trajectory whose correction is
sufficient to steer the system from failure to success.
Formally, let $\Omega(\tau) \in \{0, 1\}$ be a binary evaluation function indicating failure ($\Omega(\tau)=0$) or success ($\Omega(\tau)=1$), and $\mathcal{R}(\tau, t, a'_t)$ be an oracle rectification operator, which represents an idealized process where the original action $a_t$ is replaced by a theoretically perfect, oracle-provided action $a'_t$, and all subsequent steps are re-simulated. Naturally, the set of all decisive agent-step pairs $\mathcal{C}(\tau)$ can be expressed as:
\begin{equation}\label{eq:error-pairs}
\mathcal{C}(\tau) = \left\{ (i, t) \ \middle| \ i=\mu(t) \text{ and } \Omega(\tau)=1 \land \Omega\left(\mathcal{R}(\tau, t, a'_t)\right)=0 \right\},
\end{equation}
from which we select the root cause by targeting the earliest error in time. The definitive failure-responsible agent $i^*$ and decisive error step $t^*$ are therefore given by:
\begin{equation}\label{eq:t_star}
(i^*, t^*) = \underset{(i, t) \in \mathcal{C}(\tau)}{\arg\min} \ t.
\vspace{-0.2em}
\end{equation}
Consequently, the goal of our failure tracer, \ourmethod, is to take a failed trajectory $\tau$ as input and output the failure-responsible agent $i^*$ and the decisive error step $t^*$.

\vspace{-0.5em}
\section{Methodology}
\vspace{-0.7em}

\begin{figure*}[!t]
\centering
\includegraphics[width=\linewidth]{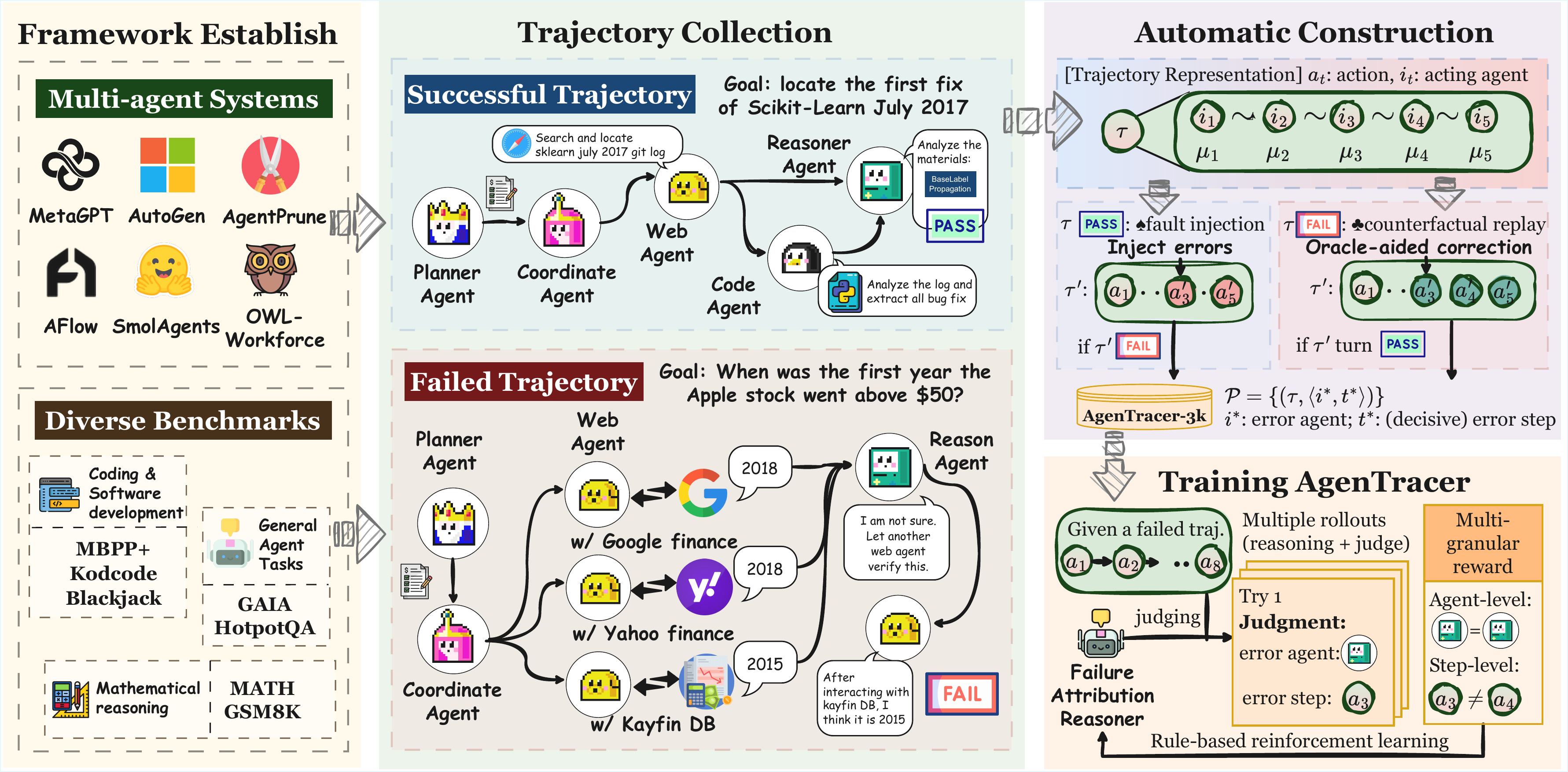}
\vspace{-1.2em}
\caption{The overview of our proposed \ourmethod.}
\label{fig:framework}
\vspace{-1.3em}
\end{figure*}

\Cref{fig:framework} illustrates the pipeline of \ourmethod and the training process of \ourmodel. Specifically, \ourmethod aggregates trajectories from six mainstream multi-agent frameworks and six datasets, which then applies programmatic fault injection (for successful trajectories) and counterfactual corrections (for failed ones) to identify the decisive error step, yielding 2,000+ trajectory–error step pairs, collectively referred to as \ourdata-2.5K. We then train a dedicated failure attributor, \ourmodel, using RL guided by multi-grained rewards to accurately locate errors.

\vspace{-0.5em}
\subsection{\ourmethod: Automatic Trajectory Annotation}
\vspace{-0.5em}

\paragraph{Collection.}
\ourmethod begin by considering a collection of $M (=6)$ distinct multi-agent systems, $\{ \mathcal{M}_m \}_{m=1}^{M}$, and a corresponding set of queries for each system, $\mathcal{D}_m = \{ \mathcal{Q}_{j}^{(m)} \}_{j=1}^{n_m}$, drawn from various task domains. For each query $\mathcal{Q}_{j}^{(m)}$, $\mathcal{M}_m$ executes and generates a raw trajectory $\tau_{j}^{(m)}$. For the specific frameworks and datasets used, please refer to \Cref{sec:exp_steup}. The trajectories from each system can be split into two sets, namely the successful ($\mathcal{T}_{\text{succ}}^{(m)}$) and failure ($\mathcal{T}_{\text{fail}}^{(m)}$) corpus:
\begin{equation}
\mathcal{T}_{\text{succ}}^{(m)} = \{ \tau_j^{(m)} \mid \Omega(\tau_j^{(m)}) = 1 \}, \quad \text{and} \quad
\mathcal{T}_{\text{fail}}^{(m)} = \{ \tau_j^{(m)} \mid \Omega(\tau_j^{(m)}) = 0\}.
\end{equation}
The overall successful and failure set can be expressed as $\mathcal{T}_\text{succ} = \bigcup_{m=1}^{M} \mathcal{T}_{\text{succ}}^{(m)}$ and $\mathcal{T}_\text{fail} = \bigcup_{m=1}^{M} \mathcal{T}_{\text{fail}}^{(m)}$, respectively, which serve as the raw input for the subsequent annotation stages.

\paragraph{Locating Decisive Errors.}
To empirically identify the decisive error pair $(i^*, t^*)$ for each failed trajectory $\tau \in \mathcal{T}_\text{fail}$, we now detail our practical implementation of the rectification operator $\mathcal{R}$ introduced in \Cref{eq:error-pairs}, which is is orchestrated by an analyzer agent $\pi_{\text{analyzer}}$ to conduct counterfactual intervention. $\pi_{\text{analyzer}}$ is provided with the full context of a failure: the trajectory $\tau$, environmental feedback $\mathcal{F}$ (\textit{e.g.}, code execution errors, tool errors), and the ground-truth solution $\mathcal{G}$ for $\mathcal{Q}$. For each step $t$ in the trajectory, the analyzer proposes a minimally invasive, corrected action $a'_t$ designed to rectify the local error without revealing the complete solution:
\begin{equation}\label{eq:analyzer}
    a'_t \leftarrow \pi_{\text{analyzer}}(s_t, a_t, \mathcal{H}_t, \mathcal{F}, \mathcal{G}).
\end{equation}
By sequentially applying this analyzer-driven intervention for $t\in\{0, 1, \ldots, T\}$ and evaluating the outcome $\Omega(\mathcal{R}(\tau, t, a'_t))$, we can systematically search for the earliest step $t^*$ that satisfies the condition in \Cref{eq:t_star}. The agent active at this step, $i^* = \mu(t^*)$, is labeled as the problematic agent, yielding a precise annotation $(\tau, \langle i^*, t^*\rangle)$ for our dataset. This process (as presented in \Cref{algo:line:part1begin,algo:line:part1-1,algo:line:part1-2,algo:line:part1-3,algo:line:part1-4,algo:line:part1-5,algo:line:part1-6,algo:line:part1-7,algo:line:part1end} of \Cref{algo:annotation_pipeline}), applied across the entire $\mathcal{T}_\text{fail}$, forms $\mathcal{D}^{-} = \{ (\tau, \langle i^*, t^* \rangle) \mid \tau \in \mathcal{T}_\text{fail} \}$.

\vspace{-0.7em}
\paragraph{Utilizing Successful Trajectories.}
To further augment our dataset with high-precision annotations, we also leverage the corpus $\mathcal{T}_\text{succ}$ through \textbf{programmatic fault injection}. Its core principle is to take a known-good trajectory and programmatically introduce a fault, thereby creating a synthetic failure instance where the decisive error is known by construction. Specifically, for each successful trajectory $\tau \in \mathcal{T}_\text{succ}$, we select a step $t$ to serve as the injection point, at which a perturbation operator $\Pi$ is utilized to the original action $a_t$ to generate a corrupted action $\tilde{a}_t = \Pi(a_t)$. A new, synthetically-generated trajectory $\tilde{\tau}$ is created by substituting this corrupted action:
\begin{equation}
    \tilde{\tau} = \mathcal{R}(\tau, t, \tilde{a}_t).
\end{equation}
If this injection process successfully induces a failure (\textit{i.e.}, $\Omega(\tilde{\tau}) = 0$), the pair $\langle\mu(t), t\rangle$ is, by definition, the decisive error for $\tilde{\tau}$. This allows us to generate a set of positive-sample datasets $\mathcal{D}^{+}$:
\begin{equation}\label{eq:positive-set}
    \mathcal{D}^{+} = \left\{ (\tilde{\tau}, \langle i^*, t^* \rangle) \mid \tau \in \mathcal{T}_{\text{succ}}, \ \tilde{\tau} = \mathcal{R}(\tau, t, \Pi(a_t)), \ \Omega(\tilde{\tau})=0, \ \langle i^*, t^* \rangle = \langle\mu(t), t\rangle \right\}.
\end{equation}
The overall process is also elaborated in \Cref{algo:line:part2-1,algo:line:part2-2,algo:line:part2-3,algo:line:part2-4,algo:line:part2-5,algo:line:part2-6,algo:line:part2-7,algo:line:part2-8,algo:line:part2-9,algo:line:part2end} of \Cref{algo:annotation_pipeline}.
\vspace{-0.5em}
\paragraph{Curated Dataset.} 
By uniting the annotations derived from both failed trajectories and synthetically generated ones, we construct our final, comprehensive dataset, denoted as $\mathcal{D}_{\text{tracer}} = \mathcal{D}^{-} \cup \mathcal{D}^{+}$, 
which we also refer to as \ourdata-2.5K, comprising over 2,000 high-fidelity annotated trajectory-error pairs. Detailed statistics on the dataset's composition are placed in \Cref{app:dataset_stats}.

\begin{algorithm}[!t]\small
\DontPrintSemicolon
\SetAlgoLined
\LinesNumbered
\KwIn{Corpus of failed trajectories $\mathcal{T}_\text{fail}$; Corpus of successful trajectories $\mathcal{T}_\text{succ}$; Analyzer agent $\pi_\text{analyzer}$; perturbation operator $\Pi$; Rectification operator $\mathcal{R}$}
\KwOut{The curated dataset $\mathcal{D}_{\text{tracer}}$}

% % \tcp{Initialize empty annotation sets}
Initialize dataset: $\mathcal{D}^- \leftarrow \emptyset$, $\mathcal{D}^+ \leftarrow \emptyset$ \;
\label{algo:line:part1begin}
\tcc{Part 1: Locate Decisive Errors via Counterfactual Intervention}
\label{algo:line:part1-1}
\For{\rm{each trajectory } $\tau \in \mathcal{T}_\text{fail}$}{
\label{algo:line:part1-2}
    \For{$t \leftarrow 0$ \KwTo $T$}{
    \label{algo:line:part1-3}
        $a'_t \leftarrow \pi_{\text{analyzer}}(s_t, a_t, \mathcal{H}_t, \mathcal{F}, \mathcal{G})$\tcp{Analyzer agent proposes a corrected action} 
        \label{algo:line:part1-4}
        $\tau' \leftarrow \mathcal{R}(\tau, t, a'_t)$\tcp{Apply intervention to generate a new trajectory} 
        \label{algo:line:part1-5}
        \If{$\Omega(\tau') = 1$}{
        \label{algo:line:part1-6}
            $i^* \leftarrow \mu(t)$,  $t^* \leftarrow t$, 
            $\mathcal{D}^- \leftarrow \mathcal{D}^- \cup \{(\tau, \langle i^*, t^* \rangle)\}$ \;
            \label{algo:line:part1-7}
            \textbf{break} \tcp{Found the earliest decisive error}
        }
    }
}
\label{algo:line:part1end}
\tcc{Part 2: Generate Errors via Programmatic Fault Injection}
\label{algo:line:part2-1}
\For{\rm{each trajectory } $\tau \in \mathcal{T}_\text{succ}$}{
\label{algo:line:part2-2}
    \text{Let } $\mathcal{J} \leftarrow \{0, 1, \ldots, T\}$ \text{ be the set of possible injection points} \;
    \label{algo:line:part2-3}
    \text{Let } $\mathcal{J}_{\text{sample}} \leftarrow \texttt{RandomSample}(\mathcal{J}, K)$ \tcp{Randomly select $K$ steps}
    \label{algo:line:part2-4}
    \For{$t \in \mathcal{J}_{\text{sample}}$}{
        \label{algo:line:part2-5}
        $\tilde{a}_t \leftarrow \Pi(a_t)$ \tcp{Perturb original action to create a fault}
        \label{algo:line:part2-6}       
        $\tilde{\tau} \leftarrow \mathcal{R}(\tau, t, \tilde{a}_t)$  \tcp{Apply injection to generate a new trajectory}
        \label{algo:line:part2-7}
        \If{$\Omega(\tilde{\tau}) = 0$}{
            \label{algo:line:part2-8}
            $i^* \leftarrow \mu(t)$, $t^* \leftarrow t$,
            $\mathcal{D}^+ \leftarrow \mathcal{D}^+ \cup \{(\tilde{\tau}, \langle i^*, t^* \rangle)\}$ \;
            \label{algo:line:part2-9}
            \textbf{break} \tcp{Stop after first successful injection for this trajectory}
        }
    }
}
\label{algo:line:part2end}
% \tcp{Combine both sets to form the final dataset}
%  \;
\Return curated dataset $\mathcal{D}_{\text{tracer}} \coloneqq \mathcal{D}^- \cup \mathcal{D}^+$
\caption{Automated Trajectory Annotation Pipeline of \ourmethod}
\label{algo:annotation_pipeline}
% \vspace{-0.3em}
\end{algorithm}

\vspace{-0.3em}
\subsection{\ourmodel: Training Agentic Failure Tracers}
\label{sec:ourmodel}
\vspace{-0.5em}
Having curated the $\mathcal{D}_{\text{tracer}}$ dataset, we proceed to train our failure tracer, \ourmodel, whose base model is set as \textsc{Qwen3-8B}. This phase aims to incentivize the model's ability to accurately pinpoint decisive errors within complex, long-horizon trajectories. Since our approach is orthogonal to the RL algorithm, we conduct the experiments based on a widely used online RL method, Group Relative Policy Optimization (GRPO)~\citep{guo2025deepseek-r1}.

\vspace{-0.5em}
\paragraph{Online Reinforcement Learning.}
 For each trajectory $\tau$ sampled from $\mathcal{D}_{\text{tracer}}$, the current policy $\pi_{\text{old}}$ generates a group of $G$ candidate decisive error pairs, $\{ \langle \hat{i}_k, \hat{t}_k \rangle \}_{k=1}^G$, each of which is evaluated against the ground-truth annotation $\langle i^*, t^* \rangle$ using a multi-granular reward function $R_k$, which will be detailed below. Unlike the standard GRPO, we omit the KL divergence term and introduce a dynamic clipping parameter $B_s$, which has been demonstrated to better balance exploration and exploitation throughout training~\citep{liu2025guardreasoner-vl}. The RL objective is thus formulated as:
\begin{equation}\label{eq:grpo}
    \mathcal{L}_{\text{RL}} = -\mathbb{E}_{\tau, \{\hat{p}_k\}_{k=1}^G} \left[ \frac{1}{G} \sum_{k=1}^G \min(\rho_k A_k, \text{clip}(\rho_k, 1-B_s, 1+B_s)A_k) \right],
\end{equation}
where $\hat{p}_k = \langle \hat{i}_k, \hat{t}_k \rangle$, the policy ratio is $\rho_k = \frac{\pi_{\text{tracer}}(\hat{p}_k|\tau)}{\pi_{\text{old}}(\hat{p}_k|\tau)}$, the estimated advantage is $A_k = (R_k - \text{mean}(\{R_j\})) / (\text{std}(\{R_j\}) + \epsilon)$ with $\epsilon=\num{1e-6}$ being a small constant. The dynamic clipping parameter $B_s$ is defined as $B_s = \max(0.2\cdot B_0, \, B_0(1-\frac{s}{S_{\text{total}}}))$, with $s$ as the current training step and $S_{\text{total}}$ the total number of steps. This dynamic schedule intuitively encourages broader exploration in the initial stages of training and gradually shifts to more stable exploitation as the policy converges.

\vspace{-0.4em}
\paragraph{Multi-Granular Reward Design.}
Regarding the implementation of advantage estimation in \Cref{eq:grpo}, we introduce a multi-granular reward designed to evaluate both the correctness of the attribution and the structural integrity of the output. The total reward $R_k$ for a candidate prediction $\hat{p}_k$ is a gated combination of content accuracy and format compliance:
\begin{equation}\label{eq:overall-loss}
    R(\hat{p}_k) = \mathbb{I}_{\text{format}} \cdot \left( \lambda \cdot r_{\text{step}}(\hat{t}_k) + (1-\lambda) \cdot r_{\text{agent}}(\hat{i}_k) \right),
\end{equation}
whose components are further defined as follows:
\vspace{-0.4em}
\begin{itemize}[leftmargin=1em,itemsep=-0.1em]
    \item[$\blacktriangleright$] \textbf{Format Reward} $\mathbb{I}_{\text{format}}$ is a strict binary reward that equals $1$ if and only if the model's output adheres to the required structure: reasoning must be enclosed within $\mathsf{\langle think\rangle\cdots\langle /think\rangle }$ tags, followed by a final answer within $\mathsf{\langle answer\rangle\cdots\langle /answer\rangle }$ tags. Furthermore, the answer itself must be formatted as $\mathsf{\langle agentID\rangle \;|\;\langle stepID \rangle}$ for accurate extraction. 

    \item[$\blacktriangleright$] \textbf{Agent-Level Reward} $r_{\text{agent}}$  is a coarse-grained, binary reward that measures whether the tracer correctly identifies the failure-responsible agent $i^*$, defined as $  r_{\text{agent}}(\hat{i}_k) = \mathbb{I}(\hat{i}_k = i^*)$, where $\mathbb{I}(c)$ is a binary indicator for the accuracy of located problematic agent. 

    \item[$\blacktriangleright$] \textbf{Step-Level Reward} $r_{\text{step}}$ incentivizes temporal proximity to the true decisive error $t^*$. We use a Gaussian kernel where the reward decays smoothly as the predicted step $\hat{t}_k$ moves away from $t^*$:
    \begin{equation}\label{eq:step-loss}
        r_{\text{step}}(\hat{t}_k) = \exp\left(-\frac{(\hat{t}_k - t^*)^2}{2\sigma^2}\right),
    \end{equation}
    where $\sigma$ controls how sharply the reward penalizes distance from the correct step.
\end{itemize}

\vspace{-0.3em}
This multi-granular design creates a smoother reward landscape for failure localization, as the partial credit from $r_{\text{step}}$ stabilizes training. Simultaneously, the hard gating from $r_{\text{format}}$ ensures the model produces reliably parsable outputs. Through online RL with these designs, we obtain a reasoning-based multi-agent failure attributor \ourmodel.

% \clearpage

\begin{table}[!tbp]
\centering
\small
\caption{Performance comparison on the Who\&When benchmark. For each subset, evaluation is conducted at both the agent and step levels. Each cell reports two values: the left corresponds to the setting \textit{w/ $\mathcal{G}$} (the failure tracer has access to ground truth trajectory), and the right corresponds to \textit{w/o $\mathcal{G}$}. The best and second-best results are \textbf{bolded} and \underline{underlined}, respectively.}
\vspace{-0.9em}
\setlength{\tabcolsep}{14pt}
\label{tab:who_when}
\begin{tabular}{l cc cc}
 \Xhline{1.2pt}
& \multicolumn{2}{c}{\textbf{Who\&When (handcraft)}} & \multicolumn{2}{c}{\textbf{Who\&When (automated)}} \\
\cmidrule(lr){2-3} \cmidrule(lr){4-5}
\multirow{-2}{*}{\makecell{\textbf{Model}}} & \makecell{\small Agent-level} & \makecell{\small Step-level} & \makecell{\small Agent-level} & \makecell{\small Step-level} \\
 \Xhline{0.5pt}
\rowcolor{CadetBlue!20}\textsc{Qwen3-8B} & 42.10/39.50 & 1.72/3.45 & 58.73/60.32 & 3.97/5.56 \\
\textsc{Llama-3.2-3B} & 37.93/50.00 & 1.72/3.45 & 37.30/45.23 & 2.38/8.73 \\
% \rowcolor{CadetBlue!20}\textsc{Qwen-2.5-7B} & & & & \\
 \Xhline{0.5pt}

\rowcolor{CadetBlue!20}\textsc{Qwen3-32B} & 44.80/44.80 & 1.72/1.72 & 63.49/57.93 & 9.52/8.73  \\
% \rowcolor{CadetBlue!20}\textsc{Qwen3-Coder} & 51.72/53.45 & 1.72/1.72 & 30.95/36.50  & 4.76/6.35 \\
\textsc{Qwen3-Coder} & 51.72/\underline{60.35} & 8.62/13.79 & 42.86/36.50  & 34.13/32.54 \\

 \Xhline{0.5pt}
% \textsc{GPT-4o} & & & & \\
\rowcolor{CadetBlue!20}\textsc{GPT-4.1} & 43.10/37.93 & 3.44/3.44 & 55.55/59.52 & 29.52/21.90 \\
\textsc{DeepSeek-R1} & \underline{56.90}/53.44 & 13.29/6.90 & \underline{66.67}/\textbf{65.08} & 31.32/29.52 \\
\rowcolor{CadetBlue!20}\textsc{Gemini-2.5-pro} & 51.72/51.72 & 9.72/6.90 & 61.11/57.14 & 29.52/25.86 \\
% \textsc{Claude-Sonnet-4} & 50.00/50.00 & 1.72/5.17 & 63.49/58.73 & 8.73/9.52 \\
\textsc{Claude-Sonnet-4} & \underline{56.90}/50.00 & \underline{17.24}/\underline{18.97} & 57.93/51.11 & \underline{40.65}/\textbf{38.83} \\

 \Xhline{0.5pt}
\rowcolor{CadetBlue!20}\ourmethod & \textbf{69.10}/\textbf{63.82} & \textbf{20.68}/\textbf{20.68} & \textbf{69.62}/\underline{63.73} & \textbf{42.86}/\underline{37.30}  \\
 \Xhline{1.2pt}
\end{tabular}
\vspace{-1em}
\end{table}

\vspace{-0.8em}
\section{Experiments}
\vspace{-0.8em}
\subsection{Experimental Setup}\label{sec:exp_steup}
\vspace{-0.5em}
\paragraph{Dataset Curation.} For collecting \ourdata-2.5K, we opt for six widely-used multi-agent systems, comprehensively incorporating all automation levels: \ding{110} \textbf{manually configured}, including MetaGPT~\citep{meta-gpt}, AutoGen~\citep{autogen} and Smolagents\footnote{\url{https://github.com/huggingface/smolagents}}; \ding{110} \textbf{partially automated}, including AgentPrune~\citep{zhang2024cut}; \ding{110} \textbf{fully automated}, including AFlow~\citep{zhang_aflow_2024} and OWL-Workforce~\citep{hu2025owloptimizedworkforcelearning}. Six benchmarks from three domains include \ding{110} \textbf{coding}, MBPP+~\citep{liu2023codegeneratedchatgptreally}, KodCode~\citep{xu2025kodcodediversechallengingverifiable} and Blackjack~\citep{meta-gpt}; \ding{110} \textbf{general agentic tasks}, GAIA~\citep{mialon2023gaia}; \ding{110} \textbf{mathematical reasoning}, MATH~\citep{hendrycksmath2021} and GSM8K~\citep{arXiv2021_Verifier-Math}. 

\vspace{-0.9em}
\paragraph{Environment.} All experimental results are obtained on one server with 8 NVIDIA H100 (80 GB)
GPUs. For RL training in \Cref{sec:ourmodel}, we use the verl\footnote{\url{https://github.com/volcengine/verl}} training platform. 

\vspace{-0.9em}
\paragraph{Model \& Parameter Configuration.} The analyzer agent $\pi_\text{analyzer}$ in \Cref{eq:analyzer} and perturbation operator $\Pi$ in \Cref{eq:positive-set} are both based on \textsc{DeepSeek-R1}~\citep{guo2025deepseek-r1} (see prompts in \Cref{app:prompt}). The coefficient $\lambda$ in \Cref{eq:overall-loss} is consistently set as $0.5$, and the parameter $\sigma$ in \Cref{eq:step-loss} equals $1$. The LLM backbone used for training \ourmodel is \textsc{Qwen3-8B}. For RL traning in \Cref{sec:ourmodel}, we set batch size to \num{32},  rollout number \num{8}, and learning rate \num{1e-6}.
%More training parameters are specified in \Cref{app:exp:hyperparameter}.

\vspace{-0.9em}
\paragraph{Benchmarks \& Evaluation.} To evaluate the failure attribution capability of \ourmodel, we adopt the Who\&When benchmark~\citep{zhang2025agent-who-and-when}, which comprises two subsets: a handcrafted set derived from Magnetic-One~\citep{fourney2024magenticonegeneralistmultiagentsolving}, and an automated set constructed from AG2~\citep{githubGitHubAg2aiag2}. Both subsets provide unseen trajectories with respect to \ourmodel. In addition, we sample a held-out test split from \ourdata-2.5K using a 9:1 ratio, yielding three evaluation subsets (devided by domains): \ourdata-code, \ourdata-math, and \ourdata-agentic. Detailed statistics are reported in \Cref{app:dataset_detail}. For evaluation, following~\citep{zhang2025agent-who-and-when}, we adopt two primary metrics: \underline{\textit{agent-level accuracy}} and \underline{\textit{step-level accuracy}}. The former measures whether the attributor correctly identifies the faulty agent $i^*$ within a trajectory, while the latter assesses whether the specific erroneous step $t^*$ is localized. We consider two evaluation settings: \textit{(i) w/ $\mathcal{G}$}, where the attributor has access to the ground truth $\mathcal{G}$ during failure attribution, and \textit{(i) w/o $\mathcal{G}$} without such access. The latter setting is harder and particularly valuable. We follow the ``all-at-once'' setting introduced in MAST, where the entire trajectory is provided to the LLM in a single pass, as \cite{zhang2025agent-who-and-when} has demonstrated this to be the most stable and effective.

\vspace{-0.7em}
\paragraph{Baselines.} We compare \ourmethod agaist LLM baselines of varying scales, encompassing \textbf{small-size models} such as \textsc{Qwen3-8B}~\citep{yang2025qwen3technicalreport} and \textsc{Llama-3.2-3B}~\citep{grattafiori2024llama}; \textbf{medium-size models}, including \textsc{Qwen3-32B} and \textsc{Qwen3-Coder-480B-A35B-Instruct} (\textsc{Qwen3-Coder})~\citep{yang2025qwen3technicalreport}; and \textbf{large-size models}, which primarily consist of state-of-the-art LLMs, such as \textsc{GPT-4.1}~\citep{gpt4.1}, \textsc{Gemini-2.5-Pro}~\citep{comanici2025gemini}, \textsc{Claude-4-Sonnet}~\citep{anthropicClaudeSonnet} and also \textsc{DeepSeek-R1}~\citep{guo2025deepseek-r1}.

\begin{table}[!tbp]
\centering
\small
\caption{Performance comparison on different subsets of \ourdata. For each subset, accuracy is reported at the agent/step levels. Each cell reports two values: the left corresponds to the setting \textit{w/ $\mathcal{G}$}, and the right \textit{w/o $\mathcal{G}$}. The best and second-best results are \textbf{bolded} and \underline{underlined}, respectively.}
\setlength{\tabcolsep}{4pt}
\vspace{-0.8em}
\label{tab:tracer-bench}
\begin{tabular}{l cc cc cc}
 \Xhline{1.2pt}
& \multicolumn{2}{c}{\textbf{Code}} & \multicolumn{2}{c}{\textbf{MATH}} & \multicolumn{2}{c}{\textbf{Agentic}} \\
\cmidrule(lr){2-3} \cmidrule(lr){4-5} \cmidrule(lr){6-7}
\multirow{-2}{*}{\makecell{\textbf{Model}}} & \makecell{\small Agent-level} & \makecell{\small Step-level} & \makecell{\small Agent-level} & \makecell{\small Step-level} & \makecell{\small Agent-level} & \makecell{\small Step-level} \\
 \Xhline{0.5pt}
\rowcolor{CadetBlue!20}\textsc{Qwen3-8B} & 45.35/32.99 & 2.36/1.15 & 31.74/33.58 & 9.52/12.96 & 27.93/30.16 & 13.49/15.31 \\
\textsc{Llama-3.2-3B} & 13.38/11.81 & 2.36/3.93 & 15.87/14.28 & 4.76/3.17 & 8.11/13.15 & 2.18/5.09 \\
% \rowcolor{CadetBlue!20}\textsc{Qwen-2.5-7B} & & & & & & \\
 \Xhline{0.5pt}
\rowcolor{CadetBlue!20}\textsc{Qwen3-32B} & 62.99/63.78 & 2.36/1.75 & 17.46/17.46 & 4.76/7.93 & 30.55/30.55 & 18.60/15.31 \\
\textsc{Qwen3-Coder} & 69.29/\underline{66.92} & 14.96/\underline{14.17} & 28.57/33.58 & 11.11/22.22 & 40.67/43.12 & 24.99/25.69 \\
% \textsc{GPT-4o} & & & & & & \\
 \Xhline{0.5pt}
\rowcolor{CadetBlue!20}\textsc{GPT-4.1} & 59.84/53.54 & 12.59/11.02 & 39.55/35.18 & 35.81/26.63 & 43.61/40.67 & 24.99/25.69 \\
\textsc{DeepSeek-R1} & 11.81/11.81 & 10.23/10.23 & 42.68/38.58 & 29.52/18.51 & 45.12/46.19 & 27.13/21.80 \\
\rowcolor{CadetBlue!20}\textsc{Gemini-2.5-pro} & \underline{70.07}/\underline{66.92} & 11.02/6.29 & \underline{58.79}/\underline{58.79} & 32.22/27.40 & 37.16/32.18 &  18.60/17.04 \\
\textsc{Claude-Sonnet-4} & 65.98/63.78 & 15.51/11.02 & 46.03/50.79 & \underline{38.10}/\underline{46.03}  &  \textbf{55.20}/\underline{49.13} & \underline{30.33}/\underline{29.80} \\
% \textsc{Claude-Sonnet-4}& &  & 46.03/50.79 & 38.10/46.03 &  & \\
 \Xhline{0.5pt}
\rowcolor{CadetBlue!20}\ourmethod & \textbf{72.95}/\textbf{72.21} & \textbf{18.85}/\textbf{18.85} & \textbf{59.32}/\textbf{66.10} & \textbf{57.63}/\textbf{57.63} & \underline{53.28}/\textbf{50.61} & \textbf{36.17}/\textbf{35.55} \\
 \Xhline{1.2pt}
\end{tabular}
\end{table}

\begin{figure*}[!t]
\centering
\vspace{-0.8em}
\includegraphics[width=\linewidth]{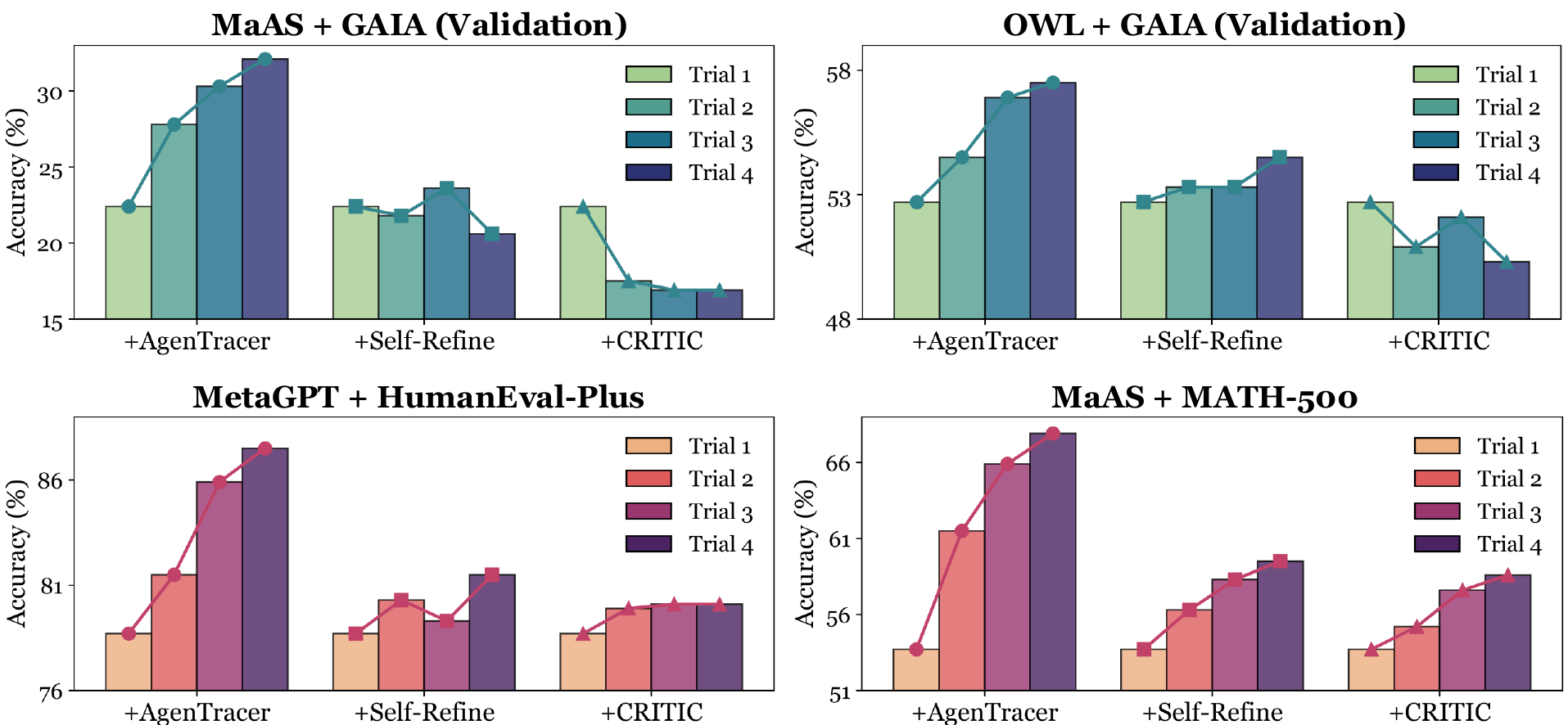}
\vspace{-1.5em}
\caption{The multi-turn improvement performance brought by \ourmodel compared with classical agent reflection baselines, Self-Refine, and CRITIC.}
\label{fig:improve}
\vspace{-1.3em}
\end{figure*}

\vspace{-0.6em}
\subsection{Main Results}\label{sec:main_result}
\vspace{-0.4em}
This section provides empirical evidence that \textit{\ourmodel outperforms substantially larger models in failure attribution within complex agentic systems.} \Cref{tab:who_when,tab:tracer-bench} report results on {Who\&When} and \ourdata subsets, respectively, presenting both agent-level and step-level attribution accuracy. Each table entry is divided into \textit{w/ ground truth $\mathcal{G}$} and \textit{w/o $\mathcal{G}$} during attribution.

\textbf{Observation \ding{182}: prevailing models are inadequate as failure attributors.} As shown in \Cref{tab:who_when}, smaller models such as \textsc{Qwen3-8B} and \textsc{Llama-3.2-3B} fail to deliver meaningful judgments, with step-level accuracy on Who\&When (handcrafted) remaining below $10\%$. Even substantially larger models like \textsc{DeepSeek-R1} and \textsc{GPT-4.1} perform unsatisfactorily, achieving only $31.32\%$ and $29.52\%$ step-level accuracy on Who\&When (automated) despite access to ground-truth $\mathcal{G}$. Notably, providing $\mathcal{G}$ does not consistently improve attribution accuracy; for example, on \ourdata-math, \textsc{Claude-4-Sonnet} attains $46.03\%$ (\textit{w/ $\mathcal{G}$}) versus $50.79\%$ (\textit{w/o $\mathcal{G}$}), and on Who\&When (handcrafted), \textsc{Qwen3-Coder} achieves $51.72\%$ versus $60.35\%$. This suggests that ground-truth supervision may sometimes mislead the attribution process, an observation consistent with prior findings in MAST~\citep{zhang2025agent-who-and-when}.

\textbf{Observation \ding{182}: \ourmethod consistently surpasses giant proprietary LLMs such as \textsc{Claude-4-Sonnet} in both agent- and step-level attribution.} Under the \textit{w/ $\mathcal{G}$} setting, as shown in \Cref{tab:who_when}, \ourmodel outperforms \textsc{GPT-4.1} and \textsc{Claude-4-Sonnet} on Who\&When (handcrafted) by $26.0\%$ and $12.2\%$ in agent-level accuracy, respectively. A similar trend is observed on \ourdata (\Cref{tab:tracer-bench}), where \ourmodel improves step-level accuracy on \ourdata-agentic by $22.68\%$ over its backbone \textsc{Qwen3-8B}, while also surpassing \textsc{DeepSeek-R1} ($+9.04\%$) and \textsc{Gemeni-2.5-Pro} ($+17.57\%$). More importantly, in the \textit{w/o $\mathcal{G}$} setting (arguably the more realistic scenario where ground truth is unavailable), \ourmodel remains robust: on \ourdata-math, \textsc{DeepSeek-R1} suffers a $9.21\%$ drop without $\mathcal{G}$, whereas \ourmodel maintains $57.63\%$. This strongly substantiates the real-world deployability and practical significance of \ourmodel.

\vspace{-0.3em}
\subsection{Boosting Mainstream MAS}\label{sec:combine_mas}
\vspace{-0.6em}
Having established the accuracy of \ourmethod in failure attribution, a natural question arises: \textit{what practical value does it provide?} The most direct answer is its potential to supply actionable feedback to failing LLM-based agentic systems, thereby enabling swift self-improvement. To assess this capability, we compare \ourmodel with two classical self-refinement approaches, Self-Refine~\citep{NeurIPS2023_Self-Refine} and CRITIC~\citep{gou2024critic}. Specifically, when an agentic system $\mathcal{M}$ completes a problem-solving episode and produces a failed trajectory $\tau$, we supply $\tau$ (\textit{w/o} $\mathcal{G}$) to either \ourmodel or Self-Refine/CRITIC. Each method then generates reflective feedback on the failure (for \ourmodel, this corresponds to the reasoning trace extracted from $\mathsf{\langle think \rangle \cdots \langle /think \rangle}$). This feedback is subsequently injected into $\mathcal{M}$ during the next round of problem solving, with the aim of leveraging external critique to enhance its performance. We iterate this process for three rounds, and both Self-Refine and CRITIC are instantiated using \textsc{GPT-4.1}.

\textbf{Observation \ding{184}: \ourmodel enables performance gains of up to $14\%$ for existing agentic systems.}  
To evaluate whether \ourmethod can provide beneficial feedback to both \textit{seen} and \textit{unseen} agentic systems and datasets, we consider three representative systems, MaAS~\citep{zhang2025maas}, OWL Workforce, and MetaGPT, together with GAIA, HumanEval+~\citep{liu2023codegeneratedchatgptreally}, and MATH-500 benchmarks. As shown in \Cref{fig:improve}, conventional reflection-based approaches fail to deliver meaningful insights for complex agentic trajectories. Even when powered by \textsc{GPT-4.1}, CRITIC consistently degrades performance (\textit{e.g.}, CRITIC+MaAS+GAIA accuracy drops by $-4.9\%$ at iteration-2 and $-5.5\%$ at iteration-3). Conversly, \ourmethod steadily improves outcomes across all settings. Notably, OWL is the open-source SOTA on GAIA for June 2025, yet \ourmethod still manages to boost its performance by $+4.8\%$. On MaAS+MATH-500, the gains are even more striking, reaching $+14.21\%$, substantially surpassing both Self-Refine and CRITIC. Overall, these results demonstrate that \ourmethod provides reliable corrective feedback and substantial performance improvements across diverse domains for complex agentic systems.

\vspace{-1.0em}
\subsection{Case Study}\label{sec:case-study}
\vspace{-0.7em}

\begin{figure*}[!t]
\centering
\vspace{-0.8em}
\includegraphics[width=\linewidth]{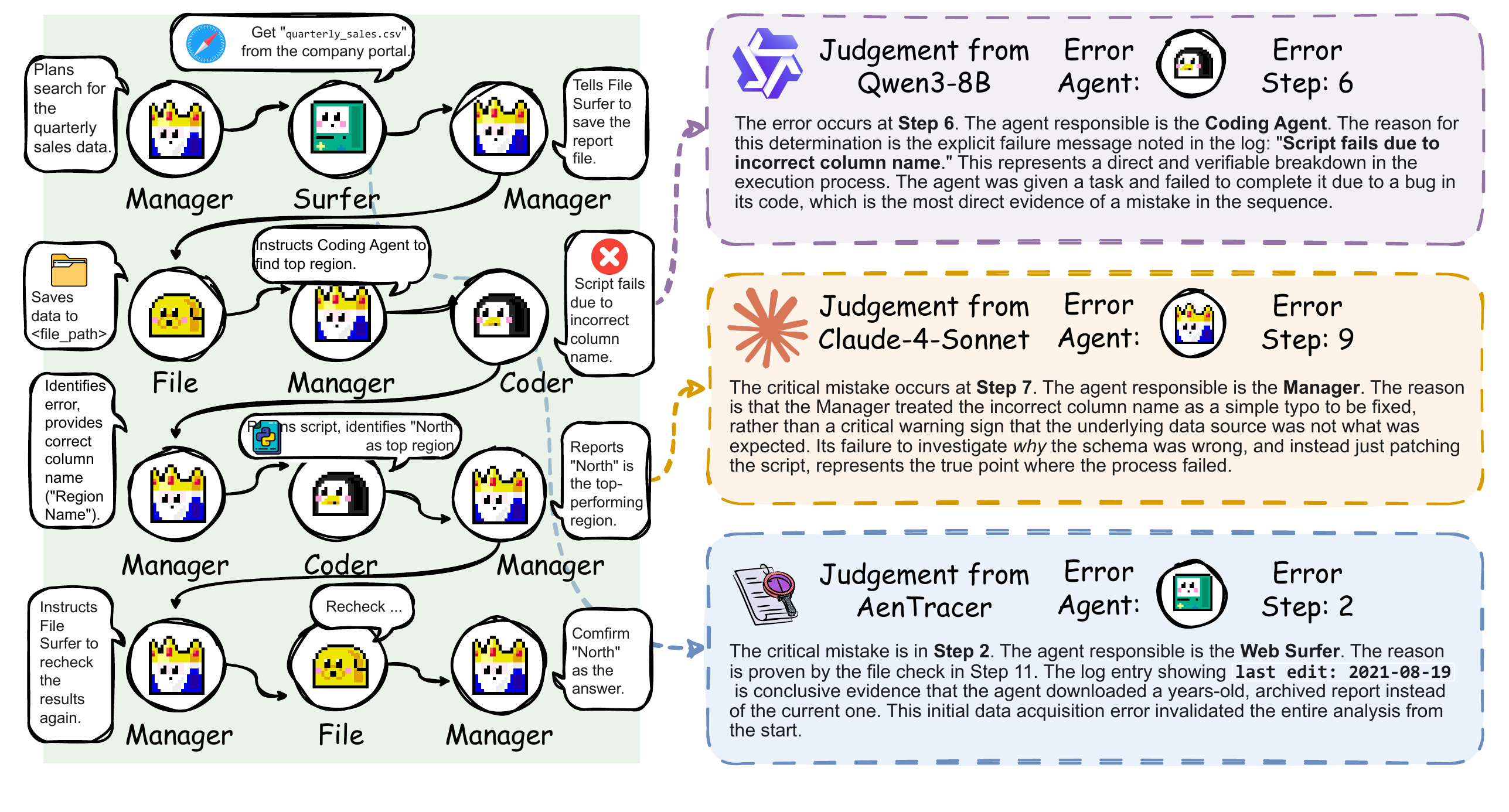}
\vspace{-2.6em}
\caption{Case study of failure attribution in a long-chain document analysis task, comparing three models (\textsc{Qwen3-8B}, \textsc{Claude-4-Sonnet}, and \ourmodel).}
\label{fig:case}
\vspace{-2.em}
\end{figure*}

\Cref{fig:case} presents a comparative case study where \textsc{Qwen3-8B}, \textsc{Claude-4-Sonnet}, and \ourmodel analyze the same failed trajectory. The task requires identifying the region with the highest infant formula sales in a company’s Q1 2024 sales data. The final (incorrect) answer produced was ``North.'' 
\textsc{Qwen3-8B} offers only a superficial diagnosis, mistakenly attributing the failure to a code execution error by the \texttt{Coder Agent} at Step 6. \textsc{Claude-4-Sonnet} goes beyond this surface-level issue and observes that the error at Step 6 may have deeper causes. In contrast, \ourmodel precisely identifies that the root cause lies in Step 2, where the \texttt{Web Surfer} agent retrieved incorrect file with wrong date, an error that only becomes apparent when analyzing evidence at Step 11. This highlights the intrinsic difficulty of failure attribution in agentic systems: errors are often subtle, originate early, and remain hidden behind seemingly correct outputs.

\vspace{-0.8em}
\section{Conclusion}
\vspace{-0.8em}

This work establishes a principled foundation for the study of agentic system failure attribution. By introducing \ourmethod, we provide the first automated framework capable of systematically generating annotated failure trajectories, as well as \ourmodel, a lightweight yet effective failure tracer that leverages multi-granular RL to achieve prevailing diagnostic accuracy. Empirical evaluation demonstrates that \ourmodel not only surpasses state-of-the-art proprietary LLMs like \textsc{Gemeni-2.5-Pro} and \textsc{Claude-4-Sonnet} on the Who\&When benchmark but also yields consistent performance gains when deployed within real-world multi-agent frameworks. Beyond advancing the state of failure attribution, our approach paves the way for self-correcting and self-evolving agentic systems, marking a step toward more resilient and autonomous collective intelligence.

\bibliography{iclr2025_conference}
\bibliographystyle{iclr2025_conference}

\appendix
\section{Dataset Details}\label{app:dataset_detail}

\begin{table}[!tbp]
\centering
\small
\caption{\ourdata dataset statistics and test set distribution across three domains, including the associated multi-agent systems. For each domain, we list the included benchmarks, the number of curated trajectories, and the subset of trajectories annotated with error-step pairs (\ourdata-2.5K).}
\label{tab:ourdata_stats}
\begin{tabular}{lccc}
\toprule
\textbf{Metric / Domain} & \textbf{Coding} & \textbf{Mathematical Reasoning}  & \textbf{General Agentic Tasks} \\
\midrule
Benchmarks & \makecell{MBPP+\\KodCode\\Blackjack}  & \makecell{MATH\\GSM8K}& \makecell{GAIA\\HotpotQA} \\
\midrule
Multi-Agent Systems & \makecell{MetaGPT\\AutoGen\\AgentPrune} & \makecell{AgentPrune\\AFlow\\AutoGen} & \makecell{Smolagents \\OWL-Workforce} \\
\midrule
Curated Trajectories & 2,170 & 1,185 & 1,300 \\
\midrule
\ourdata-2.5K & 1288 & 630 &  558 \\
\midrule
Test Set & \makecell{147} & 63 & \makecell{56} \\
\bottomrule
\end{tabular}
\end{table}

\Cref{tab:ourdata_stats} illustrates the detailed distribution of our \ourmodel

\section{Prompt Set}\label{app:prompt}

\begin{tcolorbox}[notitle, sharp corners, breakable, colframe=Periwinkle, colback=white, 
       boxrule=3pt, boxsep=0.5pt, enhanced, 
       shadow={3pt}{-3pt}{0pt}{opacity=1,mygrey},
       title={Prompt for Analyzer Agent},]\label{box:gradient}
       {\scriptsize
\begin{lstlisting}
 prompt = f"""You are a software development team tasked with diagnosing a failed programming task. Your goal is to identify the critical error in the implementation.

Task Information:
Task ID: {task_id}
Question: {question}
Ground Truth: {ground_truth}
Model Prediction: {model_prediction}

{previous_diagnosis_info}

Original Task Execution History:
{history_str}

{history_constraint}

{validation_instruction}

Your diagnosis should be in the following JSON format:
{{
    "mistake_step": <step_number>,  // The step number where the error occurred
    "mistake_agent": "<the agent that made the mistake>",  // The agent that made the mistake (e.g., "Engineer", "Architect", "ProductManager", etc.)
    "reason": <detailed_explanation>,  // Detailed explanation of why this step is wrong
    "suggested_fix": <fix_guidance>  // Guidance on how to fix the error, NOT the complete solution
}}

Important Guidelines:
1. DO NOT provide the complete solution in the suggested_fix. Only provide guidance on how to fix the error.
2. Focus on identifying the root cause of the failure.
3. The 'mistake_step' should be a number corresponding to a step in the implementation process.
4. The 'mistake_agent' should be the specific agent that made the mistake (e.g., "Engineer", "Architect", "ProductManager", "TeamLeader", "DataAnalyst").
5. The 'reason' should be detailed and explain why the current implementation is incorrect.
6. The 'suggested_fix' should provide clear guidance without giving away the complete solution.
7. Analyze the original task execution history to understand the context and identify where things went wrong.
8. CRITICAL: Before submitting, verify that your mistake_step exists in the history and your mistake_agent matches the agent that actually performed that step.

IMPORTANT: To save the diagnosis result, you MUST use the Editor.create_file command with the following format:
First, create the file using Editor.create_file command
{{
    "command_name": "Editor.create_file",
    "args": {{
        "filename": "{task_id.replace('/', '_')}_diagnosis.json"
    }}
}}
Then use command to modify the content.
Please analyze the task and provide your diagnosis in the specified JSON format. The diagnosis result should be saved to a file named '{task_id.replace('/', '_')}_diagnosis.json' in the workspace directory."""
\end{lstlisting}
}
\end{tcolorbox}

\begin{tcolorbox}[notitle, sharp corners, breakable, colframe=Periwinkle, colback=white, 
       boxrule=3pt, boxsep=0.5pt, enhanced, 
       shadow={3pt}{-3pt}{0pt}{opacity=1,mygrey},
       title={Prompt for Analyzer Agent},]\label{box:gradient}
       {\scriptsize
\begin{lstlisting}
prompt = f"""You are an intelligent attack expert analyzing a successful execution for a programming task.

Your goal is to identify the most critical step in the execution history that, if modified, would lead to an incorrect final result.

Task Information:
Task ID: {task_id}
Question: {success_log['question']}
Ground Truth: {success_log['ground_truth']}
Model Prediction: {success_log['model_prediction']}

{previous_attack_info}

Original Task Execution History:
{history_str}

{history_constraint}

Your attack analysis should be in the following JSON format:
{{
    "attack_step": <step_number>,  // The step number where the attack should be applied
    "mistake_agent": "<the agent that made the mistake>",  // The agent that made the mistake
    "original_content": "<brief description of what this step does>",  // Brief description of the original step
    "attack_content": "<SPECIFIC, IMPLEMENTABLE attack instructions with concrete code changes>",  // SPECIFIC attack instructions
    "reason": "<detailed explanation of why this step is critical and how the attack would work>"  // Detailed explanation
}}

CRITICAL REQUIREMENTS FOR ATTACK_CONTENT:
1. The attack_content MUST be SPECIFIC and IMPLEMENTABLE
2. Include CONCRETE CODE CHANGES or MODIFICATIONS
3. Provide SPECIFIC instructions that can be directly executed
4. Focus on FUNCTIONALITY CHANGES that will cause the program to fail
5. Examples of good attack_content:
   - "Change the return statement from 'return tup + (dct,)' to 'return list(tup) + [dct]'"
   - "Modify the function to return None instead of the tuple"
   - "Add a bug: change 'return tup + (dct,)' to 'return tup + (dct, dct)' (duplicate the dictionary)"
   - "Change the function to ignore the dictionary: 'return tup'"
6. AVOID vague instructions like "return incorrect type" or "modify the function"

Important Guidelines:
1. Focus on identifying the root cause of potential failure, not just any step.
2. The 'attack_step' should be a number corresponding to a step in the implementation process.
3. The 'mistake_agent' should be the agent that made the mistake.
4. The 'original_content' should briefly describe what the step does.
5. The 'attack_content' MUST be SPECIFIC and IMPLEMENTABLE with concrete changes.
6. The 'reason' should be detailed and explain why this step is critical and how the attack would work.
7. Analyze the original task execution history to understand the context and identify where things could go wrong.
8. Focus on steps that involve code generation, implementation, or key algorithmic decisions.

CRITICAL REQUIREMENTS:
1. You MUST create the file FIRST using Editor.create_file
2. You MUST write the content SECOND using Editor.write
3. You MUST use the exact filename: "{task_id.replace('/', '_')}_attack_analysis.json"
4. You MUST NOT use the 'end' command until both file operations are completed
5. You MUST provide the attack analysis in valid JSON format

Step-by-step process:
1. First, create the file:
```json
[
    {{
        "command_name": "Editor.create_file",
        "args": {{
            "filename": "{task_id.replace('/', '_')}_attack_analysis.json"
        }}
    }}
]
```

2. Then, write the attack analysis content:
```json
[
    {{
        "command_name": "Editor.write",
        "args": {{
            "path": "{task_id.replace('/', '_')}_attack_analysis.json",
            "content": "{{"attack_step": "...", "original_content": "...", "attack_content": "SPECIFIC CODE CHANGES HERE", "reason": "..."}}"
        }}
    }}
]
```

3. Only after both file operations are successful, use the end command:
```json
[
    {{
        "command_name": "end"
    }}
]
```

Please analyze the task and provide your attack analysis."""
\end{lstlisting}
}
\end{tcolorbox}

% \section{Experiment Details}\label{app:exp}

% \subsection{Training Hyperparameters}\label{app:exp:hyperparameter}

\end{document}